\documentclass[letterpaper, 10 pt, conference]{ieeeconf}  

\IEEEoverridecommandlockouts 

\usepackage{cite}
\usepackage{amsmath,amssymb,amsfonts}
\usepackage{algorithmic}
\usepackage{graphicx}
\usepackage{textcomp}
\usepackage{xcolor}
\usepackage{caption}
\usepackage{subcaption}

\def\BibTeX{{\rm B\kern-.05em{\sc i\kern-.025em b}\kern-.08em
    T\kern-.1667em\lower.7ex\hbox{E}\kern-.125emX}}

\graphicspath{{images/}}

\begin{document}

\title{\LARGE \bf Challenges in Applying Robotics to Retail Store Management}

\author{Vartika Sengar, Aditya Kapoor, Nijil George, Vighnesh Vatsal, \\Jayavardhana Gubbi, Balamuralidhar P and Arpan Pal
\thanks{The authors are with TCS Research \& Innovation,
        Tata Consultancy Services, Bengaluru, Karnataka - 560066, India. e-mail:
         {\tt\small vartika.sengar, aditya.kapoor1, george.nijil, vighnesh.vatsal, j.gubbi, balamurali.p, arpan.pal <@tcs.com>}}
}%

\maketitle


\section{The Challenge}

\textbf{An autonomous retail store management system entails inventory tracking, store monitoring, and anomaly correction. Recent attempts at autonomous retail store management have faced challenges primarily in perception for anomaly detection, as well as new challenges arising in mobile manipulation for executing anomaly correction. Advances in each of these areas along with system integration are necessary for a scalable solution in this domain.}

\section{Detailed Challenge Description}

In a typical retail store scenario, inventory is arranged according to planograms---visual schematics for object placement on shelves.
The role of human associates is to replenish any gaps in the inventory based on the planogram, check for misplaced items, and perform object rearrangement and clean-ups.
These actions can be collectively termed as ``anomaly correction" in a retail scenario.

Recently, companies such as Simbe Robotics~\cite{Simbe} and Zippedi~\cite{Zippedi} have launched mobile robots that scan shelves, digitize inventory, and alert human associates to any anomalies. 
While these robots are able to track inventory, they still rely on the intervention of human associates to perform even elementary manipulation tasks for correcting anomalies.

An autonomous system for anomaly correction would involve a mobile robot with visual sensing similar to~\cite{Simbe}, along with manipulation capabilities using a robotic arm and gripper. 
Such a solution would offer more value to a retail establishment by automating labor-intensive human tasks while remaining within comparable levels of capital expenditure to a detection-only robot. 

The first step in the solution pipeline for such a mobile robot is the detection of anomalies through visual perception. Computer vision-based approaches have been successfully applied to the retail shelf inspection context~\cite{3} for object and anomaly detection~\cite{4}. 
Their performance can be attributed to advances in deep neural networks for visual perception~\cite{2}. 
The systems for automated shelf monitoring define this problem as simultaneous detection and recognition of objects placed on a store shelf using different types of feature descriptors such as key points, gradients, patterns, colors, and feature embeddings from deep learning~\cite{3}. These systems typically use a two-stage process to detect anomalies by checking whether or not the observed image is in agreement with the pre-defined planogram~\cite{17,18,19,20,21,22}. In the first stage, the products are detected using methods such as window-based HOG~\cite{17}, BOW~\cite{17}, template matching based on morphological gradients~\cite{21} and recurrent pattern recognition~\cite{22}. In the second stage, matching is done between the observations of the first stage and the actual layout in the planogram through techniques such as sub-graph isomorphism~\cite{20} and spectral graph matching~\cite{22}.

However, the dependence on planograms limits the usage of these systems in retail settings due to periodic layout updates and constant introduction of anomalies by human agents. 
Previous efforts by TCS for shelf inspection had employed this approach~\cite{Ray_2018_ECCV}, though there were challenges arising from accurate object detection in cluttered scenes, classification with extremely high number of classes, and adaptability towards new classes.
These factors, along with class imbalances, large label spaces, and low inter-class variance resulted in limited success of this method in practice. 

The next step in the solution pipeline is task and motion planning (TAMP) for the mobile manipulator robot. 
The challenge for planning involves extracting the key components of a transition system for formulating a plan~\cite{1} through visual perception. 
The above-mentioned vision-based methods give distributed representations whose features cannot be directly utilized by the planning module, which relies on symbols to specify goals and predicates~\cite{garrett2021integrated}. To bridge between these two paradigms, inspired by the Neuro-Symbolic Concept Learner~\cite{23}, we propose to extract concept-level representations. Generally, Neuro-Symbolic AI methods include a predefined set of concepts with their values learned based on questions and answers~\cite{23,24,25}. However, the choice of concepts, and how to handle previously unseen concepts/attributes, remains an open problem. Another challenge is that the concepts must be represented in a structured manner, and the knowledge should allow for reasoning at different levels of abstraction. We propose the use of hierarchical scene graphs for this purpose. 
Executing the task and motion plan in a closed-loop manner forms the final step of the pipeline.

\begin{figure*}[ht]
\centering
\includegraphics[width=0.9\textwidth]{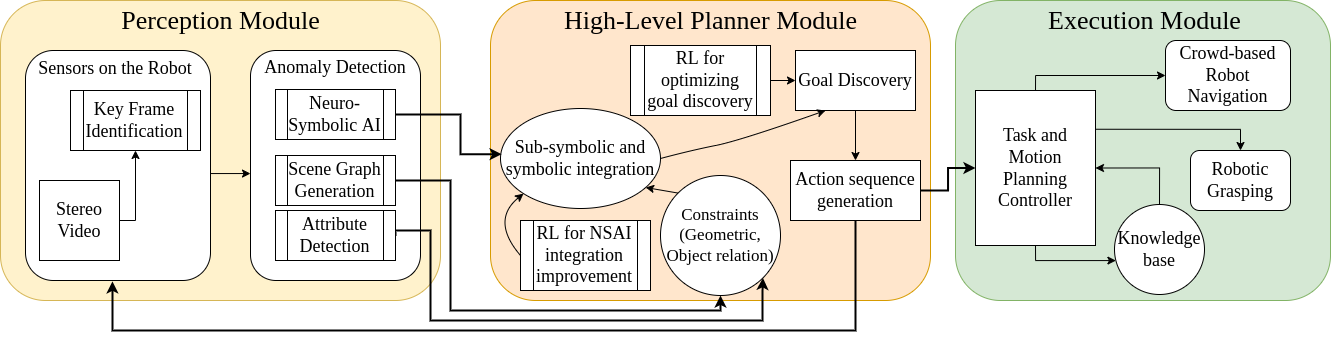}
\caption{The proposed solution pipeline for autonomous anomaly correction in a retail store.}
\label{fig:schematic}
\end{figure*}

\section{Solution}
The key idea is to develop knowledge representations from visual perception and reason about goals for a robot when presented with anomalies and unknown situations, leading to dynamic task and motion planning for mobile manipulation. This solution pipeline is illustrated in Fig.~\ref{fig:schematic}. 
As an example, if the robot 
sees an object that is placed on an incorrect shelf, the perception module would provide the present position via sensory input and target position via predefined knowledge for the object. By parsing this information into a symbolic goal, the high-level planner would generate a sequence of actions with respect to the robot's motion primitives. 
Finally, the plan would be executed by moving the mobile base, grasping the object, moving to the correct shelf and placing the object there, with feedback control at every step. 

\subsection{Neuro-Symbolic AI for Perception}

In order to avoid planogram matching-based anomaly detection, we define the problem as detection of out-of-context products, where context is based on concepts/attributes, e.g. color or shape-based anomalies. We aim to learn both local object-based concepts like location, color, size, shape, spatial relation (behind, front, left, right), texture etc. These concepts will be represented by embeddings in a latent space and are obtained from concept specific neural operators, e.g.   
$concept\: embedding\: for\: shape\; = \; ShapeOf(object\: representation)$,   
where $ShapeOf(.)$ is a neural operator corresponding to the shape concept.  
Based on the similarity measure between the object's concept embedding and embedding for each shape such as round, square, etc. the value of shape of the given object will be identified. The knowledge obtained from the proposed perception stage, i.e. the values identified corresponding to each concept will be represented in the form of a scene graph. In~\cite{24}, scene de-rendering in the form of structured scene representation is discussed. This stage will also be responsible for storing predicates representing states. For example: based on the $Relation$ concept: predicates will be- $object1 \; isBehind \; object2$; $object1\;  hasClass \; Mango$; $object1\;  hasShape\;  Round$. As described in~\cite{26}, we also need a hybrid reasoning engine to process the obtained knowledge at different levels of abstraction and use the result for task planning, e.g. `Count all the mangoes on the ground'.

\subsection{Task and Motion Planning}
The output of the perception pipeline would aid the Task and Motion Planning (TAMP) Module in achieving the desired goal in the environment using the robot. TAMP combines discrete AI approaches to task planning with continuous approaches from robotics for motion planning~\cite{5}. The problem statements for TAMP may be specified through AI planning languages such as STRIPS and PDDL.
There have been recent advancements in learning symbolic representations~\cite{7} and operators~\cite{6} for TAMP. 
Building upon previous work by TCS on structured robot re-planning using PDDL~\cite{10}, we propose to use PDDLStream~\cite{garrett2020pddlstream}, a language that combines symbolic plans with black-box sampling for continuous tasks to solve constrained optimization problems in a hybrid domain.
While these state-of-the-art planners are able to solve for TAMP problems, including long-horizon manipulation~\cite{wang2021learning}, they require the goal state to be manually specified.
The challenge in this module therefore condenses to parsing the output of the Neuro-Symbolic perception module into a goal specification for PDDLStream, through a parser or sequence-to-sequence model.
Finally, deep reinforcement learning (DRL) can be used over repeated tasks to optimise and refine the integration of perception and goal realisation, as shown in Fig.~\ref{fig:schematic}, following the bi-level framework described in~\cite{jiang2019task}.

\subsection{Manipulation and Grasping}
Once a plan is generated by the TAMP module, it needs to be executed on the physical robot.
This involves moving the robot base through environments with humans using crowd navigation~\cite{chen2019crowd}, and grasping objects on shelves in an optimal manner with feedback control to achieve closed-loop manipulation~\cite{prattichizzo2016grasping}.
The state-of-the-art vision-based systems that make use of DRL-based techniques on parallel-jaw grippers~\cite{mahler2018dex} and vacuum suction grippers~\cite{mahler2019learning} have been successfully applied for manipulation~\cite{levine2016end}. We plan to apply these strategies, as well as grasp planning in tight, constrained and cluttered spaces~\cite{lundell2021ddgc}, along with the development of soft grasping hardware and DRL-based planners~\cite{pozzi2020hand} that can handle a large selection of objects (fragile, irregularly-shaped etc.) encountered in retail stores.

\section{Conclusion}
Automating the management of a retail store in terms of anomaly correction presents unique research and engineering challenges in terms of perception and mobile manipulation. While there are state-of-the-art techniques in each of these modules, we described a potential solution for deploying them as a whole in real-world settings.  

\bibliographystyle{IEEEtran}
\bibliography{IEEEabrv,refs}

\end{document}